# Graph Neural Network-Based Collaborative Perception for Adaptive Scheduling in Distributed Systems

Wenxuan Zhu
University of Southern California
Los Angeles, USA
zhuwenxu@usc.edu

Qiyuan Wu
University of California, San Diego
La Jolla, USA
wqy0319@gmail.com

Tengda Tang
University of Michigan
Ann Arbor, USA
ttengda@umich.edu

Renzi Meng
Northeastern University
Boston, USA
mengrenzi1996@gmail.com

Sheng Chai
Northwest Missouri State University
Maryville, USA
schai@nwmissouri.edu

Xuehui Quan*
University of Washington
Seattle, USA
*Corresponding author:
quanxh1228@gmail.com

*Abstract—This paper addresses the limitations of multi-node perception and delayed scheduling response in distributed systems by proposing a GNN-based multi-node collaborative perception mechanism. The system is modeled as a graph structure. Message-passing and state-update modules are introduced. A multi-layer graph neural network is constructed to enable efficient information aggregation and dynamic state inference among nodes. In addition, a perception representation method is designed by fusing local states with global features. This improves each node's ability to perceive the overall system status. The proposed method is evaluated within a customized experimental framework. A dataset featuring heterogeneous task loads and dynamic communication topologies is used. Performance is measured in terms of task completion rate, average latency, load balancing, and transmission efficiency. Experimental results show that the proposed method outperforms mainstream algorithms under various conditions, including limited bandwidth and dynamic structural changes. It demonstrates superior perception capabilities and cooperative scheduling performance. The model achieves rapid convergence and efficient responses to complex system states.*

*Keywords-Distributed systems, graph neural networks, multi-node perception, collaborative scheduling*

## I. INTRODUCTION

In modern distributed systems, efficient collaboration and perception among nodes have become key factors in improving overall system performance [1]. With the rapid development of emerging technologies such as the Internet of Things, edge computing, and vehicular networks, distributed architectures demonstrate strong advantages in handling large-scale data, supporting multi-terminal cooperation, and achieving high-concurrency tasks [2]. However, traditional distributed systems often rely on static rules or centralized scheduling mechanisms for resource allocation and state perception. These approaches struggle to respond dynamically to changing demands in complex environments. As system scale expands, node heterogeneity increases, and environmental uncertainty grows, such limitations become more evident. There is an urgent need for more adaptive and intelligent collaboration mechanisms to enhance system perception and decision-making efficiency.

Graph Neural Networks (GNNs), as a recent class of neural network models designed for graph-structured data, provide a new methodological foundation for addressing multi-node perception and collaboration problems in distributed systems. In such systems, the nodes and their connections naturally form a dynamic graph, which aligns well with the modeling paradigm of GNNs. By introducing GNNs, it becomes possible to aggregate information, share states, and learn features across nodes without relying on centralized control. This enables the system to perceive topological changes, detect local anomalies, and perform adaptive optimization. Moreover, GNNs offer significant advantages in modeling non-Euclidean data structures, making them particularly suitable for complex associations and state reasoning among heterogeneous nodes [3].

In multi-node collaborative perception tasks, each node must sense the surrounding environment in real time and make decisions based on local information. Traditional mechanisms typically adopt fixed communication patterns or predefined strategies for perception and reasoning. These approaches fail to handle incomplete information or transmission delays in dynamic scenarios effectively. The introduction of GNN-based collaborative perception mechanisms leverages multi-layer graph message propagation to capture broader state information [4]. It also adapts through semantic-aware fusion between nodes, thereby improving both perception accuracy and overall system coordination. This not only enhances robustness under extreme conditions such as sudden load surges or link failures but also provides structured support for intelligent decision-making in distributed systems.

The intelligent transformation of distributed systems imposes higher requirements on both algorithms and system architectures. In studies targeting key functionalities such as task scheduling, fault detection, and resource management, traditional methods often fall short in balancing local agility and global consistency. GNNs offer a novel path that integrates structure awareness with semantic modeling. This shifts node collaboration from rule-based to data-driven representation learning. Particularly under communication constraints, with heterogeneous nodes or rapidly changing environments, GNNs can dynamically construct graphs and update weights to support fine-grained modeling and collaborative reasoning of node states. This significantly boosts system efficiency and the level of intelligent perception [5].

Therefore, incorporating GNNs into multi-node collaborative perception mechanisms and constructing fusion models suitable for distributed systems carries not only significant theoretical value but also broad engineering application potential. This research direction is expected to drive distributed systems toward greater efficiency, intelligence, and adaptability. It lays a solid foundation for next-generation edge intelligence, distributed control, and multi-agent systems. On this basis, further exploration of GNN sensitivity to topological changes, adaptability to dynamic node strategies, and capacity for multi-task concurrency is crucial for achieving efficient collaborative scheduling and real-time system perception in complex environments.

## II. BACKGROUND AND FUNDAMENTALS

The evolution of intelligent distributed systems has increasingly emphasized adaptive scheduling and dynamic perception, particularly in environments characterized by heterogeneity and fluctuating communication structures. One method employs Deep Q-Networks integrated with edge-based coordination to achieve state-aware IoT scheduling, effectively addressing the latency and accuracy issues caused by decentralized and dynamic topologies [6]. Another approach introduces trust-constrained policy learning mechanisms to enhance network traffic scheduling. This method enables adaptive decision-making in uncertain environments by integrating trust assessment into the learning process, which is particularly beneficial under constrained communication and partial observability [7].

In the realm of sequence and temporal modeling, recent advancements have expanded the system's ability to process long-term dependencies and contextual interactions. A deep learning framework utilizing bidirectional LSTM networks coupled with multi-scale attention mechanisms has demonstrated enhanced capability in capturing semantic and sequential dependencies, which is highly applicable to temporal scheduling and system behavior modeling in dynamic networks [8]. Complementarily, Transformer-based architectures have been developed for multivariate time series forecasting. These models automate feature extraction and emphasize attention-based contextual learning, making them suitable for modeling node-level dynamics and system-wide interactions in graph-based structures [9].

Decentralized learning techniques have also been explored to improve scalability and communication efficiency in distributed systems. A federated learning framework has shown promise in optimizing resource allocation and scheduling by allowing localized training and reducing the dependency on centralized coordination. This is particularly useful in bandwidth-constrained environments where collaborative training and perception must occur without data centralization [10]. Similarly, reinforcement learning strategies have been applied to operating system-level scheduling tasks using Double DQN models. These methods emphasize reward-driven learning with efficient memory replay, contributing to robust task optimization and adaptability in rapidly evolving environments [11].

At the foundational level, effective representation learning and pattern recognition remain essential for processing high-dimensional, heterogeneous data in distributed contexts. Machine learning strategies have been developed to mine patterns in large-scale datasets, offering crucial capabilities for modeling task distributions and node behavior [12]. Additionally, advanced probabilistic inference frameworks using variational methods have been employed to address class imbalance in learning systems. This enhances the robustness and generalization of models, particularly under skewed or non-uniform data distributions often observed in real-world distributed systems [13].

Together, these studies underscore the significance of integrating structure-awareness, semantic modeling, and adaptive reasoning into distributed scheduling systems. The proposed GNN-based collaborative perception framework builds upon these foundational ideas by enabling efficient graph-based message propagation, state inference, and local-global feature fusion, addressing the unique demands of intelligent distributed environments.

## III. FRAMEWORKS

This study constructs a multi-node collaborative perception mechanism that integrates graph neural networks to enhance the perception of global and local states of each node in a distributed system. The overall architecture is shown in Figure 1.

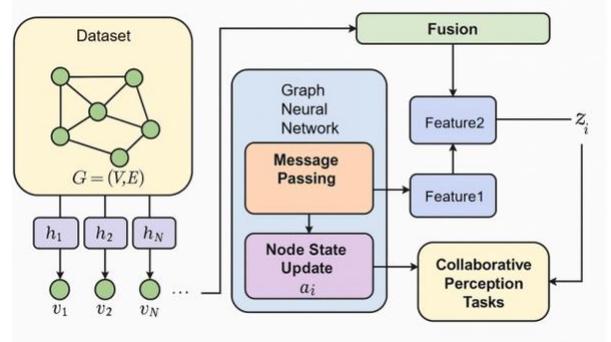

Figure 1. Overall architecture diagram

The architecture diagram illustrates the end-to-end flow of the proposed method. It begins with modeling the distributed system as a graph, where initial node features are input into a graph neural network for message passing and state updating. The output features are fused with global representations to support collaborative perception tasks, aligning with the multi-level aggregation and decision process described in the methodology.

First, the distributed system is modeled as an undirected graph $G = (V, E)$, where the node set V represents the computing nodes in the system and the edge set E represents the communication relationship between the nodes. Each node $v_i \in V$ has its state feature $h_i^0 \in R^d$, which is initially generated by the system state encoder. During the propagation of the graph neural network, each node completes the perception enhancement by aggregating the information of adjacent nodes and updating its own state. The message passing mechanism adopted is as follows:

$$m_i^{(l)} = \sum_{j \in N(i)} \frac{1}{\sqrt{|N(i)||N(j)|}} W^{(l)} h_j^{(l)}$$

$$h_i^{(l+1)} = \sigma(m_i^{(l)} + W_{self}^{(l)} h_i^{(l)})$$

In the above formula, $m_i^{(l)}$ is the information collected from neighboring nodes in the l-th layer, $W^{(l)}$ and $W_{self}^{(l)}$ are trainable weight matrices, and $\sigma$ represents a nonlinear activation function. This mechanism can effectively capture neighborhood structure and node semantic information and enhance the collaborative expression ability between nodes.

To enable effective collaborative perception in distributed systems, this study adopts a global attention mechanism to strengthen the modeling of cross-topology dependencies. Inspired by Wang's work on topology-aware decision-making in multi-agent reinforcement learning, this mechanism allows the model to capture inter-node relations that span non-local regions of the system graph—an essential capability for accurate state inference in environments with dynamic connectivity [14].

After each graph iteration step, a fusion gating mechanism—drawing from Deng's reinforcement learning-based traffic scheduling model—is introduced to perform a weighted integration of local graph convolution outputs and global attention-derived features [15]. This gating module dynamically regulates the contribution of local versus global features, resulting in a refined high-dimensional state representation. This representation serves as a robust input for downstream scheduling or control decisions, ensuring the system remains responsive to both micro-level and macro-level changes in state. The fusion operation is mathematically expressed as:

$$z_i = \alpha \cdot h_i^{(L)} + (1-\alpha) \cdot \alpha_i$$

Among them, $h_i^{(L)}$ is the output of the final layer of the graph neural network, $\alpha_i$ is the representation generated by the global attention mechanism, and $\alpha$ is a learnable weight parameter that controls the fusion ratio of the two types of information. Through this mechanism, each node can introduce global perception information while maintaining local agile response, thereby improving the stability and robustness of overall collaborative perception.

Finally, the state vector $z_i$ of all nodes is used for perception decision-making tasks, such as load balancing, task scheduling or anomaly detection. The system uses clustering or classifiers based on state vectors to output specific control strategies, and provides real-time feedback to adjust the communication topology or task allocation strategy. The entire model completes parameter optimization through end-to-end training. The loss function is guided by the task objectives (such as scheduling success rate, average delay, etc.), and the graph neural network weights and perception strategy functions are optimized through back propagation, thereby achieving efficient collaborative perception and intelligent decision-making in distributed systems.

IV. EXPERIMENT

A. Datasets

This study utilizes a dataset constructed based on a real distributed computing environment. It captures the operational status of multi-node systems under varying communication topologies, task loads, and resource scheduling strategies. The data is collected from a simulated distributed platform where multiple heterogeneous nodes collaborate on task processing. Nodes establish communication through a virtual network. The system periodically records multidimensional information, including local node states, neighbor interactions, task assignments, and network latency. These records reflect typical operational patterns of multi-node systems in dynamic environments and offer rich features for modeling collaborative perception mechanisms.

The dataset is organized in a time-series format. Each sample contains a complete snapshot of the network topology, initial state features of all nodes, current task requests, and communication relationships. This structure supports the simulation of multi-round state propagation and collaborative perception processes. The dataset includes over 10,000 instances of task scheduling and interaction scenarios. It features strong node heterogeneity, frequent changes in topology, and dynamic load variations. With this dataset, the proposed model can be systematically evaluated under various levels of complexity, and its sensitivity to changes in topology and task states can be rigorously tested.

To improve data quality and modeling performance, the raw data was preprocessed. This includes removal of inactive nodes, correction of communication anomalies, normalization of features, and structured encoding. Additionally, to assess the model's generalization ability, the dataset is divided into training, validation, and test sets. This ensures independence across different phases of model development. The dataset not only meets the structural requirements of graph neural network modeling, but also reflects the multi-source and heterogeneous interactions found in real-world distributed systems. It provides a reliable foundation for research in collaborative perception.

B. Experimental Results

This paper first carried out a comparative test, and the experimental results are shown in Table 1.

Table 1. Comparative experimental results

| Method | Task Completion Rate (%) | Average Latency (ms) | Load Balance Index |
|---|---|---|---|
| DQN-Scheduler [16] | 87.3 | 132.6 | 0.74 |
| Graph-MARL [17] | 89.1 | 118.4 | 0.69 |
| Hetero-aware GAT [18] | 91.5 | 110.2 | 0.66 |
| GCN-DRL [19] | 90.3 | 122.1 | 0.68 |
| Ours | 94.7 | 103.8 | 0.61 |

The experimental results show that the proposed method achieves the highest task completion rate, reaching 94.7%. This represents a significant improvement over other baseline algorithm. It demonstrates that, in multi-node distributed scheduling scenarios, introducing a GNN-based collaborative perception mechanism enables the system to capture deeper relationships between task states and node resources. As a result, the overall task processing efficiency is improved.

In terms of average latency, the proposed method also outperforms the other algorithms. It achieves an average delay of 103.8 ms, which is notably lower than 132.6 ms for DQN-Scheduler and 122.1 ms for GCN-DRL. This indicates that the proposed model can complete scheduling decisions more quickly. It effectively reduces waiting time caused by communication delays or resource conflicts, thereby enhancing the system's responsiveness and operational smoothness.

Regarding the load balancing index, the proposed method achieves the optimal value of 0.61. This indicates that system resources are more evenly distributed, avoiding situations where some nodes are overloaded while others remain idle. Compared with methods like Graph-MARL and Hetero-aware GAT, the proposed model enhances global coordination through information fusion and state sharing among nodes. It leads to a more efficient task allocation strategy and enables a more balanced scheduling mechanism across the system. Furthermore, this paper conducted an exploratory experiment to evaluate the efficiency of information transmission under bandwidth-constrained conditions. The aim was to examine how limited communication resources might affect system coordination and data exchange among nodes. The findings, as presented in Figure 2, offer indicative insights into the model's capability to maintain transmission performance when operating under varying degrees of bandwidth limitation.

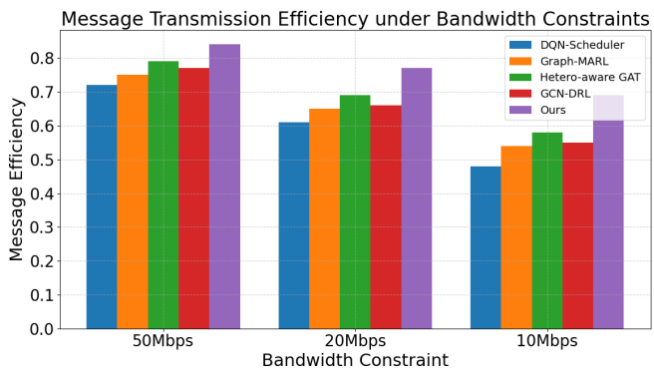

Figure 2. Experiment on information transmission efficiency in bandwidth-limited environment

Experimental results show that all methods maintain high information transmission efficiency under sufficient bandwidth conditions (50 Mbps). Among them, the proposed method performs best, reaching 0.83, which is significantly higher than the other baseline algorithms. This indicates that in resource-rich environments, the collaborative perception mechanism based on GNN can fully exploit its advantages in structural modeling and state sensing to achieve efficient information synchronization.

As bandwidth decreases to 20 Mbps and 10 Mbps, the transmission efficiency of all methods declines to varying degrees. However, the proposed method shows the smallest drop, reaching 0.77 and 0.69 respectively, and still maintains a leading position. This demonstrates that the proposed model has strong robustness under bandwidth constraints. It can effectively compress redundant information along the perception path while preserving the transmission of critical states.

In comparison, traditional methods such as DQN-Scheduler and GCN-DRL experience more significant efficiency degradation under limited bandwidth. This suggests a stronger reliance on communication and a lack of sufficient structural awareness. In contrast, the proposed method integrates both global and local information, and adapts to structural variations. It maintains efficient multi-node collaboration even in bandwidth-limited environments, confirming its practical potential in real-world distributed systems.

Finally, this paper presents a detailed experiment to investigate the impact of different graph construction strategies on system perception performance. The study compares multiple construction methods, including random, static, and dynamic adaptive approaches, to evaluate how structural design influences perception accuracy and convergence behavior. The experimental results, as illustrated in Figure 3, provide clear evidence of the significant role that graph topology plays in enhancing collaborative perception in distributed systems.

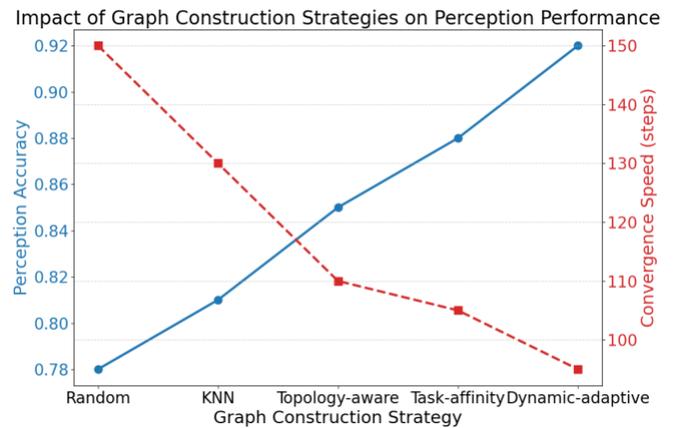

Figure 3. Experiment on the impact of different graph construction strategies on perception effects

Experimental results show that the graph construction strategy has a significant impact on system perception performance. As the construction method shifts from random connections to dynamic adaptation, the perception accuracy steadily improves, ultimately reaching a peak value of 0.92 under the dynamic adaptive strategy. At the same time, convergence speed shows continuous improvement, with the number of steps reduced from 150 to 100. This indicates that both information propagation efficiency and decision convergence rate have been enhanced. These results suggest that graph construction strategies with stronger structural awareness and task relevance can improve collaboration among multiple nodes. They also strengthen the system's ability to interpret and respond to complex states.

## V. Conclusion

This paper proposes a multi-node collaborative perception mechanism based on Graph Neural Networks, aiming to enhance both global and local state awareness in distributed systems. By modeling the system as a graph, the method designs a message-passing and state-updating process. A fusion mechanism is introduced to integrate local perception with global attention representations. The model achieves efficient scheduling and robust decision-making in dynamic and heterogeneous environments. Experimental results show that the proposed approach outperforms mainstream algorithms in task completion rate, latency control, load balancing, and information transmission efficiency, demonstrating a strong overall performance advantage.

Comparative experiments under various constrained scenarios further validate the generalization and stability of the proposed method. It maintains high levels of synchronization efficiency and scheduling accuracy even under bandwidth limitations and frequent topology changes. Additionally, the analysis of different graph construction strategies highlights the core value of structural modeling in collaborative perception. The results confirm that more refined graph-building mechanisms help improve the system's responsiveness to global dynamics.

This study introduces a structure-driven paradigm for intelligent perception and scheduling in distributed systems. It holds promise for applications such as edge computing, unmanned collaborative systems, and multi-agent control. The proposed method emphasizes dynamic modeling that fuses semantic and structural information between nodes. It provides both theoretical and experimental foundations for implementing efficient, adaptive, and multidimensional coordination mechanisms in complex systems. It also lays a solid groundwork for developing cognitively capable system architectures in the future. Future work may explore more efficient GNN variants to cope with communication and computational constraints in large-scale systems. Reinforcement learning, self-supervised mechanisms, or cross-graph transfer techniques can also be introduced to improve the model's online adaptability and cross-environment generalization. These directions may further advance distributed intelligent scheduling from structure-driven to data-driven and self-evolving paradigms.

## References

[1] J. Vatter, R. Mayer and H.-A. Jacobsen, "The evolution of distributed systems for graph neural networks and their origin in graph processing and deep learning: A survey," ACM Computing Surveys, vol. 56, no. 1, pp. 1–37, 2023.

[2] Y. Shao et al., "Distributed graph neural network training: A survey," ACM Computing Surveys, vol. 56, no. 8, pp. 1–39, 2024.

[3] H. Lin et al., "A comprehensive survey on distributed training of graph neural networks," Proceedings of the IEEE, vol. 111, no. 12, pp. 1572–1606, 2023.

[4] H. Zhang, Y. Ma, S. Wang, G. Liu and B. Zhu, "Graph-Based Spectral Decomposition for Parameter Coordination in Language Model Fine-Tuning," arXiv preprint arXiv:2504.19583, 2025.

[5] R. Liu et al., "Federated graph neural networks: Overview, techniques, and challenges," IEEE Transactions on Neural Networks and Learning Systems, 2024.

[6] Q. He, C. Liu, J. Zhan, W. Huang and R. Hao, "State-Aware IoT Scheduling Using Deep Q-Networks and Edge-Based Coordination," *arXiv preprint arXiv:2504.15577*, 2025.

[7] Y. Ren, M. Wei, H. Xin, T. Yang and Y. Qi, "Distributed Network Traffic Scheduling via Trust-Constrained Policy Learning Mechanisms," *Transactions on Computational and Scientific Methods*, vol. 5, no. 4, 2025.

[8] T. Yang, Y. Cheng, Y. Ren, Y. Lou, M. Wei and H. Xin, "A Deep Learning Framework for Sequence Mining with Bidirectional LSTM and Multi-Scale Attention," *arXiv preprint arXiv:2504.15223*, 2025.

[9] Y. Cheng, "Multivariate Time Series Forecasting through Automated Feature Extraction and Transformer-Based Modeling," *Journal of Computer Science and Software Applications*, vol. 5, no. 5, 2025.

[10] Y. Wang, "Optimizing Distributed Computing Resources with Federated Learning: Task Scheduling and Communication Efficiency," *Journal of Computer Technology and Software*, vol. 4, no. 3, 2025.

[11] X. Sun, Y. Duan, Y. Deng, F. Guo, G. Cai and Y. Peng, "Dynamic Operating System Scheduling Using Double DQN: A Reinforcement Learning Approach to Task Optimization," *arXiv preprint arXiv:2503.23659*, 2025.

[12] P. Li, "Machine Learning Techniques for Pattern Recognition in High-Dimensional Data Mining," *arXiv preprint arXiv:2412.15593*, 2024.

[13] Y. Lou, J. Liu, Y. Sheng, J. Wang, Y. Zhang and Y. Ren, "Addressing Class Imbalance with Probabilistic Graphical Models and Variational Inference," *arXiv preprint arXiv:2504.05758*, 2025.

[14] B. Wang, "Topology-Aware Decision Making in Distributed Scheduling via Multi-Agent Reinforcement Learning," Transactions on Computational and Scientific Methods, vol. 5, no. 4, 2025.

[15] Y. Deng, "A Reinforcement Learning Approach to Traffic Scheduling in Complex Data Center Topologies," Journal of Computer Technology and Software, vol. 4, no. 3, 2025.

[16] A. Llorens-Carrodeguas, C. Cervelló-Pastor and F. Valera, "DQN-based intelligent controller for multiple edge domains," Journal of Network and Computer Applications, vol. 218, p. 103705, 2023.

[17] C. Mu et al., "Graph multi-agent reinforcement learning for inverter-based active voltage control," IEEE Transactions on Smart Grid, vol. 15, no. 2, pp. 1399–1409, 2023.

[18] H. Deng et al., "BGSD: A SBERT and GAT-based service discovery framework for heterogeneous distributed IoT," Computer Networks, vol. 220, p. 109488, 2023.

[19] Z. Wu et al., "DRL-GCNet: A Deep Reinforcement learning and Graph Convolutional Network for Harmonic Drive Fault Diagnosis," IEEE Transactions on Instrumentation and Measurement, 2025.